# From Ordinary Differential Equations to Structural Causal Models: the deterministic case

**Joris M. Mooij**[*]
Institute for Computing and
Information Sciences
Radboud University Nijmegen
The Netherlands

**Dominik Janzing**
Max Planck Institute
for Intelligent Systems
Tübingen, Germany

**Bernhard Schölkopf**
Max Planck Institute
for Intelligent Systems
Tübingen, Germany

## Abstract

We show how, and under which conditions, the equilibrium states of a first-order Ordinary Differential Equation (ODE) system can be described with a deterministic Structural Causal Model (SCM). Our exposition sheds more light on the concept of causality as expressed within the framework of Structural Causal Models, especially for cyclic models.

## 1 Introduction

Over the last few decades, a comprehensive theory for acyclic causal models was developed (e.g., see (Pearl, 2000; Spirtes et al., 1993)). In particular, different, but related, approaches to causal inference and modeling have been proposed for the causally sufficient case. These approaches are based on different starting points. One approach starts from the (local or global) causal Markov condition and links observed independences to the causal graph. Another approach uses causal Bayesian networks to link a particular factorization of the joint distribution of the variables to causal semantics. The third approach uses a structural causal model (sometimes also called structural equation model or functional causal model) where each effect is expressed as a function of its direct causes and an unobserved noise variable. The relationships between these aproaches are well understood (Lauritzen, 1996; Pearl, 2000).

Over the years, several attempts have been made to extend the theory to the cyclic case, thereby enabling causal modeling of systems that involve feedback (Spirtes, 1995; Koster, 1996; Pearl and Dechter, 1996; Neal, 2000; Hyttinen et al., 2012). However, the relationships between the different approaches mentioned before do not immediately generalize to the cyclic case in general (although partial results are known for the linear case and the discrete case). Nevertheless, several algorithms (starting from different assumptions) have been proposed for inferring cyclic causal models from observational data (Richardson, 1996; Lacerda et al., 2008; Schmidt and Murphy, 2009; Itani et al., 2010; Mooij et al., 2011).

The most straightforward extension to the cyclic case seems to be offered by the structural causal model framework. Indeed, the formalism stays intact when one simply drops the acyclicity constraint. However, the question then arises how to interpret cyclic structural equations. One option is to assume an underlying discrete-time dynamical system, in which the structural equations are used as fixed point equations (Spirtes, 1995; Dash, 2005; Lacerda et al., 2008; Mooij et al., 2011; Hyttinen et al., 2012), i.e., they are used as update rules to calculate the values at time $t+1$ from the values at time $t$, and then one lets $t \to \infty$. Here we show how an alternative interpretation of structural causal models arises naturally when considering systems of ordinary differential equations. By considering how these differential equations behave in an equilibrium state, we arrive at a structural causal model that is time independent, yet where the causal semantics pertaining to interventions is still valid. As opposed to the usual interpretation as discrete-time fixed point equations, the continuous-time dynamics is not defined by the structural equations. Instead, we describe how the structural equations arise from the given dynamics. Thus it becomes evident that different dynamics can yield identical structural causal models. This interpretation sheds more light on the meaning of structural equations, and does not make any substantial distinction between the cyclic and acyclic cases.

It is sometimes argued that inferring causality amounts to simply inferring the time structure connecting the observed variables, since the cause always preceeds the effect. This, however, ignores two important facts: First, time order between two variables

---

[*]Also affiliated to the Informatics Institute, University of Amsterdam, The Netherlands

does not tell us whether the earlier one caused the later one, or whether both are due to a common cause. This paper addresses a second counter argument: a variable need not necessarily refer to a measurement performed at a certain time instance. Instead, a causal graph may formalize how intervening on some variables influences the equilibrium state of others. This describes a phenomenological level on which the original time structure between variables disappears, but causal graphs und structural equations may still be well-defined. On this level, also cyclic structural equations get a natural and well-defined meaning.

For simplicity, we consider only deterministic systems, and leave the extension to stochastic systems with possible confounding as future work.

## 2 Ordinary Differential Equations

Let $\mathcal{I} := \{1, \ldots, D\}$ be an index set of variable labels. Consider variables $X_i \in \mathcal{R}_i$ for $i \in \mathcal{I}$, where $\mathcal{R}_i \subseteq \mathbb{R}^{d_i}$. We use normal font for a single variable and boldface for a tuple of variables $\boldsymbol{X}_I \in \prod_{i \in I} \mathcal{R}_i$ with $I \subseteq \mathcal{I}$.

### 2.1 Observational system

Consider a dynamical system $\mathcal{D}$ described by $D$ coupled first-order ordinary differential equations and an initial condition $\boldsymbol{X}_0 \in \mathcal{R}_\mathcal{I}$:[1]

$$\dot{X}_i(t) = f_i(\boldsymbol{X}_{\mathrm{pa}_\mathcal{D}(i)}), \quad X_i(0) = (\boldsymbol{X}_0)_i \quad \forall i \in \mathcal{I} \quad (1)$$

Here, $\mathrm{pa}_\mathcal{D}(i) \subseteq \mathcal{I}$ is the set of (indices of) *parents*[2] of variable $X_i$, and each $f_i : \mathcal{R}_{\mathrm{pa}_\mathcal{D}(i)} \to \mathcal{R}_i$ is a (sufficiently smooth) function. This dynamical system is assumed to describe the "natural" or "observational" state of the system, without any intervention from outside. We will assume that if $j \in \mathrm{pa}_\mathcal{D}(i)$, then $f_i$ depends on $X_j$ (in other words, $f_i$ should not be constant when varying $X_j$). Slightly abusing terminology, we will henceforth call such a dynamical system $\mathcal{D}$ an Ordinary Differential Equation (ODE).

The *structure* of these differential equations can be represented as a directed graph $\mathcal{G}_\mathcal{D}$, with one node for each variable and a directed edge from $X_i$ to $X_j$ if and only if $\dot{X}_j$ depends on $X_i$.

#### 2.1.1 Example: the Lotka-Volterra model

The Lotka-Volterra model (Murray, 2002) is a well-known model from population biology, modeling the mutual influence of the abundance of prey $X_1 \in [0, \infty)$ (e.g., rabbits) and the abundance of predators $X_2 \in$

---
[1] We write $\dot{X} := \frac{dX}{dt}$.
[2] Note that $X_i$ can be a parent of itself.

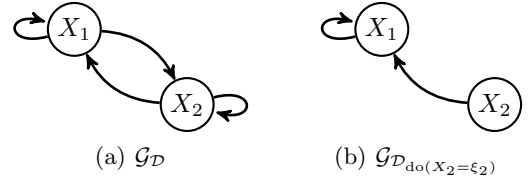

(a) $\mathcal{G}_\mathcal{D}$  (b) $\mathcal{G}_{\mathcal{D}_{\mathrm{do}(X_2 = \xi_2)}}$

Figure 1: (a) Graph of the Lotka-Volterra model (2); (b) Graph of the same ODE after the intervention $\mathrm{do}(X_2 = \xi_2)$, corresponding with (5).

$[0, \infty)$ (e.g., wolves):

$$\begin{cases} \dot{X}_1 &= X_1(\theta_{11} - \theta_{12}X_2) \\ \dot{X}_2 &= -X_2(\theta_{22} - \theta_{21}X_1) \end{cases} \quad \begin{cases} X_1(0) = a \\ X_2(0) = b \end{cases} \quad (2)$$

with all parameters $\theta_{ij} > 0$ and initial condition satisfying $a \geq 0, b \geq 0$. The graph of this system is depicted in Figure 1(a).

### 2.2 Intervened system

*Interventions* on the system $\mathcal{D}$ described in (1) can be modeled in different ways. Here we will focus on *"perfect" interventions*: for a subset $I \subseteq \mathcal{I}$ of components, we force the value of $\boldsymbol{X}_I$ to attain some value $\boldsymbol{\xi}_I \in \mathcal{R}_I$. In particular, we will assume that the intervention is active from $t = 0$ to $t = \infty$, and that its value $\boldsymbol{\xi}_I$ does not change over time. Inspired by the do-operator introduced by Pearl (2000), we will denote this type of intervention as $\mathrm{do}(\boldsymbol{X}_I = \boldsymbol{\xi}_I)$.

On the level of the ODE, there are many ways of realizing a given perfect intervention. One possible way is to add terms of the form $\kappa(\xi_i - X_i)$ (with $\kappa > 0$) to the expression for $\dot{X}_i$, for all $i \in I$:

$$\dot{X}_i(t) = \begin{cases} f_i(\boldsymbol{X}_{\mathrm{pa}_\mathcal{D}(i)}) + \kappa(\xi_i - X_i) & i \in I \\ f_i(\boldsymbol{X}_{\mathrm{pa}_\mathcal{D}(i)}) & i \in \mathcal{I} \setminus I, \end{cases} \quad (3)$$
$$X_i(0) = (\boldsymbol{X}_0)_i$$

This would correspond to extending the system by components that monitor the values of $\{X_i\}_{i \in I}$ and exert negative feedback if they deviate from their target values $\{\xi_i\}_{i \in I}$. Subsequently, we let $\kappa \to \infty$ to consider the idealized situation in which the intervention completely overrides the other mechanisms that normally determine the value of $\boldsymbol{X}_I$. Under suitable regularity conditions, we can let $\kappa \to \infty$ and obtain the *intervened system* $\mathcal{D}_{\mathrm{do}(\boldsymbol{X}_I = \boldsymbol{\xi}_I)}$:

$$\dot{X}_i(t) = \begin{cases} 0 & i \in I \\ f_i(\boldsymbol{X}_{\mathrm{pa}_\mathcal{D}(i)}) & i \in \mathcal{I} \setminus I, \end{cases}$$
$$X_i(0) = \begin{cases} \xi_i & i \in I \\ (\boldsymbol{X}_0)_i & i \in \mathcal{I} \setminus I \end{cases} \quad (4)$$

A perfect intervention changes the graph $\mathcal{G}_\mathcal{D}$ associated to the ODE $\mathcal{D}$ by removing the incoming arrows on the nodes corresponding to the intervened variables $\{X_i\}_{i\in I}$. It also changes the parent sets of intervened variables: for each $i \in I$, $\text{pa}_\mathcal{D}(i)$ is replaced by $\text{pa}_{\mathcal{D}_{\text{do}(\boldsymbol{X}_I=\boldsymbol{\xi}_I)}}(i) = \emptyset$.

### 2.2.1 Example: Lotka-Volterra model

Let us return to the example in section 2.1.1. In this context, consider the perfect intervention $\text{do}(X_2 = \xi_2)$. This intervention could be realized by monitoring the abundance of wolves very precisely and making sure that the number equals the target value $\xi_2$ at all time (for example, by killing an excess of wolves and introducing new wolves from some reservoir of wolves). This leads to the following intervened ODE:

$$\begin{cases} \dot{X}_1 &= X_1(\theta_{11} - \theta_{12}X_2) \\ \dot{X}_2 &= 0 \end{cases} \quad \begin{cases} X_1(0) = a \\ X_2(0) = \xi_2 \end{cases} \quad (5)$$

The corresponding intervened graph is illustrated in Figure 1(b).

## 2.3 Stability

An important concept in our context is *stability*, defined as follows:

**Definition 1** *The ODE $\mathcal{D}$ specified in (1) is called* stable *if there exists a unique equilibrium state $\boldsymbol{X}^* \in \mathcal{R}_\mathcal{I}$ such that for any initial state $\boldsymbol{X}_0 \in \mathcal{R}_\mathcal{I}$, the system converges to this equilibrium state as $t \to \infty$:*

$$\exists !_{\boldsymbol{X}^* \in \mathcal{R}_\mathcal{I}} \, \forall_{\boldsymbol{X}_0 \in \mathcal{R}_\mathcal{I}} : \lim_{t\to\infty} \boldsymbol{X}(t) = \boldsymbol{X}^*.$$

One can weaken the stability condition by demanding convergence to and uniqueness of the equilibrium only for a certain subset of all initial states. For clarity of exposition, we will use this strong stability condition.

We can extend this concept of stability by considering a certain set of perfect interventions:

**Definition 2** *Let $\mathcal{J} \subseteq \mathcal{P}(\mathcal{I})$.[3] The ODE $\mathcal{D}$ specified in (1) is called* stable with respect to $\mathcal{J}$ *if for all $I \in \mathcal{J}$ and for all $\boldsymbol{\xi}_I \in \mathcal{R}_I$, the intervened ODE $\mathcal{D}_{\text{do}(\boldsymbol{X}_I=\boldsymbol{\xi}_I)}$ has a unique equilibrium state $\boldsymbol{X}^*_{\text{do}(\boldsymbol{X}_I=\boldsymbol{\xi}_I)} \in \mathcal{R}_\mathcal{I}$ such that for any initial state $\boldsymbol{X}_0 \in \mathcal{R}_\mathcal{I}$ with $(\boldsymbol{X}_0)_I = \boldsymbol{\xi}_I$, the system converges to this equilibrium as $t \to \infty$:*

$$\exists!_{\boldsymbol{X}^*_{\text{do}(\boldsymbol{X}_I=\boldsymbol{\xi}_I)} \in \mathcal{R}_\mathcal{I}} \, \forall_{\substack{\boldsymbol{X}_0 \in \mathcal{R}_\mathcal{I} \, s.t. \\ (\boldsymbol{X}_0)_I = \boldsymbol{\xi}_I}} : \lim_{t\to\infty} \boldsymbol{X}(t) = \boldsymbol{X}^*_{\text{do}(\boldsymbol{X}_I=\boldsymbol{\xi}_I)}.$$

---
[3]For a set $A$, we denote with $\mathcal{P}(A)$ the power set of $A$ (the set of all subsets of $A$).

This definition can also be weakened by not demanding stability for all $\boldsymbol{\xi}_I \in \mathcal{R}_I$, but for smaller subsets instead. Again, we will use this strong condition for clarity of exposition, although in a concrete example to be discussed later (see Section 2.3.2), we will actually weaken the stability assumption along these lines.

### 2.3.1 Example: the Lotka-Volterra model

The ODE (2) of the Lotka-Volterra model is not stable, as discussed in detail by Murray (2002). Indeed, it has two equilibrium states, $(X_1^*, X_2^*) = (0, 0)$ and $(X_1^*, X_2^*) = (\theta_{22}/\theta_{21}, \theta_{11}/\theta_{12})$. The Jacobian of the dynamics is given by:

$$\nabla \boldsymbol{f}(\boldsymbol{X}) = \begin{pmatrix} \theta_{11} - \theta_{12}X_2 & -\theta_{12}X_1 \\ \theta_{21}X_2 & -\theta_{22} + \theta_{21}X_1 \end{pmatrix}$$

In the first equilibrium state, it has a positive and a negative eigenvalue ($\theta_{11}$ and $-\theta_{22}$, respectively), and hence this equilibrium is unstable. In the second equilibrium state it has two imaginary eigenvalues, $\pm i\sqrt{\theta_{11}\theta_{22}}$. One can show (Murray, 2002) that the steady state of the system is an undamped oscillation around this equilibrium.

The intervened system (5) is only generically stable, i.e., for most values of $\xi_2$: the unique stable equilibrium state is $(X_1^*, X_2^*) = (0, \xi_2)$ as long as $\theta_{11} - \theta_{12}\xi_2 \neq 0$. If $\theta_{11} - \theta_{12}\xi_2 = 0$, there exists a family of equilibria $(X_1^*, X_2^*) = (c, \xi_2)$ with $c \geq 0$.

### 2.3.2 Example: damped harmonic oscillators

The favorite toy example of physicists is a system of coupled harmonic oscillators. Consider a one-dimensional system of $D$ point masses $m_i$ ($i = 1, \ldots, D$) with positions $Q_i \in \mathbb{R}$ and momenta $P_i \in \mathbb{R}$, coupled by springs with spring constants $k_i$ and equilibrium lengths $l_i$, under influence of friction with friction coefficients $b_i$, with fixed end positions $Q = 0$ and $Q = L$ (see also Figure 2).

We first sketch the qualitative behavior: there is a unique equilibrium position where the sum of forces vanishes for every single mass. Moving one or several masses out of their equilibrium position stimulates vibrations of the entire system. Damped by friction, every mass converges to its unique and stable equilibrium position in the limit of $t \to \infty$. If one or several masses are fixed to positions different from their equilibrium points, the positions of the remaining masses still converge to unique (but different) equilibrium positions. The structural equations that we derive later will describe the change of the unconstrained equilibrium positions caused by fixing the others.

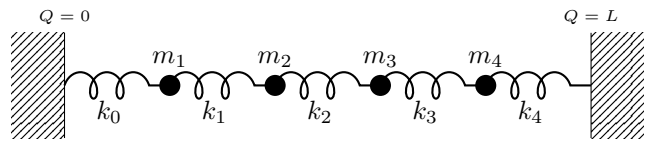

Figure 2: Mass-spring system for $D = 4$.

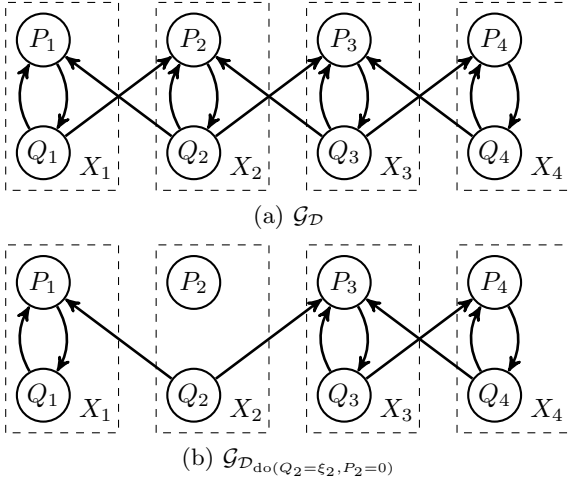

Figure 3: Graphs of the dynamics of the mass-spring system for $D = 4$. (a) Observational situation (b) Intervention $\mathrm{do}(Q_2 = \xi_2, P_2 = 0)$.

The equations of motion for this system are given by:

$$\dot{P}_i = k_i(Q_{i+1} - Q_i - l_i)$$
$$\quad - k_{i-1}(Q_i - Q_{i-1} - l_{i-1}) - \frac{b_i}{m_i}P_i$$
$$\dot{Q}_i = P_i/m_i$$

where we define $Q_0 := 0$ and $Q_{D+1} := L$. The graph of this ODE is depicted in Figure 3(a). At equilibrium (for $t \to \infty$), all momenta vanish, and the following equilibrium equations hold:

$$0 = k_i(Q_{i+1} - Q_i - l_i) - k_{i-1}(Q_i - Q_{i-1} - l_{i-1})$$
$$0 = P_i$$

which is a linear system of equations in terms of the $Q_i$. There are $D$ equations for $D$ unknowns $Q_1, \ldots, Q_D$, and one can easily check that it has a unique solution.

A perfect intervention on $Q_i$ corresponds to fixating the position of the $i$'th mass. Physically, this is achieved by adding a force that drives $Q_i$ to some fixed location, i.e., the intervention on $Q_i$ is achieved through modifying the equation of motion for $\dot{P}_i$. To deal with this example in our framework, we consider the pairs $X_i := (Q_i, P_i) \in \mathbb{R}^2$ to be the elementary variables. Consider for example the perfect intervention $\mathrm{do}(X_2 = (\xi_2, 0))$, which effectively replaces the dynamical equations $\dot{Q}_2$ and $\dot{P}_2$ by $\dot{Q}_2 = 0$, $\dot{P}_2 = 0$ and their initial conditions by $(\boldsymbol{Q}_0)_2 = \xi_2$, $(\boldsymbol{P}_0)_2 = 0$. The graph of the corresponding ODE is depicted in Figure 3(b). Because of the friction, also this intervened system converges to a unique equilibrium that does not depend on the initial value.

This holds more generally: for any perfect intervention on (any number) of pairs $X_i$ of the type $\mathrm{do}(X_i = (\xi_i, 0))$, the intervened system will converge towards a unique equilibrium because of the damping term. Interventions that result in a nonzero value for any momentum $P_i$ while the corresponding position is fixed are physically impossible, and hence will not be considered. Concluding, we have seen that the mass-spring system is stable with respect to perfect interventions on any number of position variables, which we model mathematically as a joint intervention on the corresponding pairs of position and momentum variables.

## 3 Equilibrium equations

In this section, we will study how the dynamical equations give rise to *equilibrium equations* that describe equilibrium states, and how these change under perfect interventions. This is an intermediate representation on our way to structural causal models.

### 3.1 Observational system

At equilibrium, the rate of change of any variable is zero, by definition. Therefore, an equilibrium state of the observational system $\mathcal{D}$ defined in (1) satisfies the following *equilibrium equations*:

$$0 = f_i(\boldsymbol{X}_{\mathrm{pa}_{\mathcal{D}}(i)}) \qquad \forall i \in \mathcal{I}. \quad (6)$$

This is a set of $D$ coupled equations with unknowns $X_1, \ldots, X_D$. The stability assumption (cf. Definition 1) implies that there exists a unique solution $\boldsymbol{X}^*$ of the equilibrium equations (6).

### 3.2 Intervened system

Similarly, for the intervened system $\mathcal{D}_{\mathrm{do}(\boldsymbol{X}_i = \boldsymbol{\xi}_i)}$ defined in (4), we obtain the following equilibrium equations:

$$\begin{cases} 0 = X_i - \xi_i & \forall i \in \mathcal{I} \\ 0 = f_j(\boldsymbol{X}_{\mathrm{pa}_{\mathcal{D}}(j)}) & \forall j \in \mathcal{I} \setminus I \end{cases} \quad (7)$$

If the system is stable with respect to this intervention (cf. Definition 2), then there exists a unique solution $\boldsymbol{X}^*_{\mathrm{do}(\boldsymbol{X}_I = \boldsymbol{\xi}_I)}$ of the intervened equilibrium equations (7).

Note that we can also go directly from the equilibrium equations (6) of the observational system $\mathcal{D}$ to the equilibrium equations (7) of the intervened system

$\mathcal{D}_{\mathrm{do}(\boldsymbol{X}_I=\boldsymbol{\xi}_I)}$, simply by replacing the equilibrium equations "$0 = f_i(\boldsymbol{X}_{\mathrm{pa}_\mathcal{D}(i)})$" for $i \in I$ by equations of the form "$0 = X_i - \xi_i$". Indeed, note that the modified dynamical equation

$$\dot{X}_i = f_i(\boldsymbol{X}_{\mathrm{pa}_\mathcal{D}(i)}) + \kappa(\xi_i - X_i)$$

yields an equilibrium equation of the form

$$0 = f_i(\boldsymbol{X}_{\mathrm{pa}_\mathcal{D}(i)}) + \kappa(\xi_i - X_i)$$

which, in the limit $\kappa \to \infty$, reduces to $0 = X_i - \xi_i$. This seemingly trivial observation will turn out to be quite important.

### 3.3 Labeling equilibrium equations

If we would consider the equilibrium equations as a set of *unlabeled* equations $\{\mathcal{E}_i : i \in \mathcal{I}\}$, where $\mathcal{E}_i$ denotes the equilibrium equation "$0 = f_i(\boldsymbol{X}_{\mathrm{pa}_\mathcal{D}(i)})$" (or "$0 = X_i - \xi_i$" after an intervention) for $i \in \mathcal{I}$, then we will not be able to correctly predict the result of interventions, as we do not know *which* of the equilibrium equations should be changed in order to model the particular intervention. This information is present in the dynamical system $\mathcal{D}$ (indeed, the terms "$\dot{X}_i$" in the l.h.s. of the dynamical equations in (1) indicate the targets of the intervention), but is lost when considering the corresponding equilibrium equations (6) as an unlabeled set (because the terms "$\dot{X}_i$" have all been replaced by zeroes).

This important information can be preserved by labeling the equilibrium equations. Indeed, the *labeled* set of equilibrium equations $\mathcal{E} := \{(i, \mathcal{E}_i) : i \in \mathcal{I}\}$ contains all information needed to predict how equilibrium states change on arbitrary (perfect) interventions. Under an intervention $\mathrm{do}(\boldsymbol{X}_I = \boldsymbol{\xi}_I)$, the equilibrium equations are changed as follows: for each intervened component $i \in I$, the equilibrium equation $\mathcal{E}_i$ is replaced by the equation $\tilde{\mathcal{E}}_i$ defined as "$0 = X_i - \xi_i$", whereas the other equilibrium equations $\mathcal{E}_j$ for $j \in \mathcal{I} \setminus I$ do not change. If the dynamical system is stable with respect to this intervention, this modified system of equilibrium equations describes the new equilibrium obtained under the intervention. We conclude that the information about the values of equilibrium states and how these change under perfect interventions is encoded in the labeled equilibrium equations.

### 3.4 Labeled equilibrium equations

The previous considerations motivate the following formal definition of a system of Labeled Equilibrium Equations (LEE) and its semantics under interventions.

**Definition 3** *A system of* Labeled Equilibrium Equations (LEE) $\mathcal{E}$ *for $D$ variables $\{X_i\}_{i \in \mathcal{I}}$ with $\mathcal{I} := \{1, \ldots, D\}$ consists of $D$ labeled equations of the form*

$$\mathcal{E}_i : \quad 0 = g_i(\boldsymbol{X}_{\mathrm{pa}_\mathcal{E}(i)}), \qquad i \in \mathcal{I}, \qquad (8)$$

*where $\mathrm{pa}_\mathcal{E}(i) \subseteq \mathcal{I}$ is the set of (indices of) parents of variable $X_i$, and each $g_i : \mathcal{R}_{\mathrm{pa}_\mathcal{E}(i)} \to \mathcal{R}_i$ is a function.*

The *structure* of an LEE $\mathcal{E}$ can be represented as a directed graph $\mathcal{G}_\mathcal{E}$, with one node for each variable and a directed edge from $X_i$ to $X_j$ (with $j \neq i$) if and only if $\mathcal{E}_i$ depends on $X_j$.

A perfect intervention transforms an LEE into another (intervened) LEE:

**Definition 4** *Let $I \subseteq \mathcal{I}$ and $\boldsymbol{\xi}_I \in \mathcal{R}_I$. For the perfect intervention $\mathrm{do}(\boldsymbol{X}_I = \boldsymbol{\xi}_I)$ that forces the variables $\boldsymbol{X}_I$ to take the value $\boldsymbol{\xi}_I$, the intervened LEE $\mathcal{E}_{\mathrm{do}(\boldsymbol{X}_I=\boldsymbol{\xi}_I)}$ is obtained by replacing the labeled equations of the original LEE $\mathcal{E}$ by the following modified labeled equations:*

$$0 = \begin{cases} X_i - \xi_i & i \in I \\ g_i(\boldsymbol{X}_{\mathrm{pa}_\mathcal{E}(i)}) & i \in \mathcal{I} \setminus I. \end{cases} \qquad (9)$$

We define the concept of solvability for LEEs that mirrors the definition of stability for ODEs:

**Definition 5** *An LEE $\mathcal{E}$ is called* solvable *if there exists a unique solution $\boldsymbol{X}^*$ to the system of (labeled) equations $\{\mathcal{E}_i\}$. An LEE $\mathcal{E}$ is called* solvable with respect to $\mathcal{J} \subseteq \mathcal{P}(\mathcal{I})$ *if for all $I \in \mathcal{J}$ and for all $\boldsymbol{\xi}_I \in \mathcal{R}_I$, the intervened LEE $\mathcal{E}_{\mathrm{do}(\boldsymbol{X}_I=\boldsymbol{\xi}_I)}$ is solvable.*

As we saw in the previous section, an ODE induces an LEE in a straightforward way. The graph $\mathcal{G}_{\mathcal{E}_\mathcal{D}}$ of the induced LEE $\mathcal{E}_\mathcal{D}$ is equal to the graph $\mathcal{G}_\mathcal{D}$ of the ODE $\mathcal{D}$. It is immediate that if the ODE $\mathcal{D}$ is stable, then the induced LEE $\mathcal{E}_\mathcal{D}$ is solvable. As we saw at the end of Section 3.2, our ways of modeling interventions on ODEs and on LEEs are compatible. We will spell out this important result in detail.

**Theorem 1** *Let $\mathcal{D}$ be an ODE, $I \subseteq \mathcal{I}$ and $\boldsymbol{\xi}_I \in \mathcal{R}_I$. (i) Applying the perfect intervention $\mathrm{do}(\boldsymbol{X}_I = \boldsymbol{\xi}_I)$ to the induced LEE $\mathcal{E}_\mathcal{D}$ gives the same result as constructing the LEE corresponding to the intervened ODE $\mathcal{D}_{\mathrm{do}(\boldsymbol{X}_I=\boldsymbol{\xi}_I)}$:*

$$(\mathcal{E}_\mathcal{D})_{\mathrm{do}(\boldsymbol{X}_I=\boldsymbol{\xi}_I)} = \mathcal{E}_{\mathcal{D}_{\mathrm{do}(\boldsymbol{X}_I=\boldsymbol{\xi}_I)}}.$$

*(ii) Stability of the intervened ODE $\mathcal{D}_{\mathrm{do}(\boldsymbol{X}_I=\boldsymbol{\xi}_I)}$ implies solvability of the induced intervened LEE $\mathcal{E}_{\mathcal{D}_{\mathrm{do}(\boldsymbol{X}_I=\boldsymbol{\xi}_I)}}$, and the corresponding equilibrium and solution $\boldsymbol{X}^*_{\mathrm{do}(\boldsymbol{X}_I=\boldsymbol{\xi}_I)}$ are identical.* □

### 3.4.1 Example: damped harmonic oscillators

Consider again the example of the damped, coupled harmonic oscillators of section 2.3.2. The labeled equilibrium equations are given explicitly by:

$$\mathcal{E}_i : \quad \begin{cases} 0 &= k_i(Q_{i+1} - Q_i - l_i) \\ &\quad - k_{i-1}(Q_i - Q_{i-1} - l_{i-1}) \\ 0 &= P_i \end{cases} \quad (10)$$

## 4 Structural Causal Models

In this section we will show how an LEE representation can be mapped to the more popular representation of Structural Causal Models, also known as Structural Equation Models (Bollen, 1989). We follow the terminology of Pearl (2000), but consider here only the subclass of *deterministic* SCMs.

### 4.1 Observational system

The following definition is a special case of the more general definition in (Pearl, 2000, Section 1.4.1):

**Definition 6** *A* deterministic *Structural Causal Model (SCM) $\mathcal{M}$ on $D$ variables $\{X_i\}_{i \in \mathcal{I}}$ with $\mathcal{I} := \{1, \ldots, D\}$ consists of $D$ structural equations of the form*

$$X_i = h_i(\boldsymbol{X}_{\mathrm{pa}_\mathcal{M}(i)}), \qquad i \in \mathcal{I}, \quad (11)$$

*where $\mathrm{pa}_\mathcal{M}(i) \subseteq \mathcal{I} \setminus \{i\}$ is the set of (indices of) parents of variable $X_i$, and each $h_i : \mathcal{R}_{\mathrm{pa}_\mathcal{M}(i)} \to \mathcal{R}_i$ is a function.*

Each structural equation contains a function $h_i$ that depends on the components of $\boldsymbol{X}$ in $\mathrm{pa}_\mathcal{M}(i)$. We think of the parents $\mathrm{pa}_\mathcal{M}(i)$ as the *direct causes* of $X_i$ (relative to $\boldsymbol{X}_\mathcal{I}$) and the function $h_i$ as the *causal mechanism* that maps the direct causes to the effect $X_i$. Note that the l.h.s. of a structural equation by definition contains only $X_i$, and that the r.h.s. is a function of variables *excluding* $X_i$ itself. In other words, $X_i$ is not considered to be a direct cause of itself. The *structure* of an SCM $\mathcal{M}$ is often represented as a directed graph $\mathcal{G}_\mathcal{M}$, with one node for each variable and a directed edge from $X_i$ to $X_j$ (with $j \neq i$) if and only if $h_i$ depends on $X_j$. Note that this graph does not contain "self-loops" (edges pointing from a node to itself), by definition.

### 4.2 Intervened system

A Structural Causal Model $\mathcal{M}$ comes with a specific semantics for modeling perfect interventions (Pearl, 2000):

**Definition 7** *Let $I \subseteq \mathcal{I}$ and $\boldsymbol{\xi}_I \in \mathcal{R}_I$. For the perfect intervention $\mathrm{do}(\boldsymbol{X}_I = \boldsymbol{\xi}_I)$ that forces the variables $\boldsymbol{X}_I$ to take the value $\boldsymbol{\xi}_I$, the intervened SCM $\mathcal{M}_{\mathrm{do}(\boldsymbol{X}_I = \boldsymbol{\xi}_I)}$ is obtained by replacing the structural equations of the original SCM $\mathcal{M}$ by the following modified structural equations:*

$$X_i = \begin{cases} \xi_i & i \in I \\ h_i(\boldsymbol{X}_{\mathrm{pa}_\mathcal{M}(i)}) & i \in \mathcal{I} \setminus I. \end{cases} \quad (12)$$

The reason that the equations in a SCM are called "structural equations" (instead of simply "equations") is that they also contain information for modeling interventions, in a similar way as the labeled equilibrium equations contain this information. In particular, the l.h.s. of the structural equations indicate the targets of an intervention.[4]

### 4.3 Solvability

Similarly to our definition for LEEs, we define:

**Definition 8** *An SCM $\mathcal{M}$ is called* solvable *if there exists a unique solution $\boldsymbol{X}^*$ to the system of structural equations. An SCM $\mathcal{M}$ is called* solvable with respect to *$\mathcal{J} \subseteq \mathcal{P}(\mathcal{I})$ if for all $I \in \mathcal{J}$ and for all $\boldsymbol{\xi}_I \in \mathcal{R}_I$, the intervened SCM $\mathcal{M}_{\mathrm{do}(\boldsymbol{X}_I = \boldsymbol{\xi}_I)}$ is solvable.*

Note that each (deterministic) SCM $\mathcal{M}$ with acyclic graph $\mathcal{G}_\mathcal{M}$ is solvable, even with respect to the set of all possible intervention targets, $\mathcal{P}(\mathcal{I})$. This is not necessarily true if directed cycles are present.

### 4.4 From labeled equilibrium equations to deterministic SCMs

Finally, we will now show that under certain stability assumptions on an ODE $\mathcal{D}$, we can represent the information about (intervened) equilibrium states that is contained in the corresponding set of labeled equilibrium equations $\mathcal{E}_\mathcal{D}$ as an SCM $\mathcal{M}_{\mathcal{E}_\mathcal{D}}$.

First, given an LEE $\mathcal{E}$, we will construct an induced SCM $\mathcal{M}_\mathcal{E}$, provided certain solvability conditions hold:

**Definition 9** *If the LEE $\mathcal{E}$ is solvable with respect to $\{\mathcal{I} \setminus \{i\}\}_{i \in \mathcal{I}}$, it is called* structurally solvable.

If the LEE $\mathcal{E}$ is structurally solvable, we can proceed as follows. Let $i \in \mathcal{I}$ and write $I_i := \mathcal{I} \setminus$ $\{i\}$. We define the induced parent set $\mathrm{pa}_{\mathcal{M}_\mathcal{E}}(i) := \mathrm{pa}_\mathcal{E}(i) \setminus \{i\}$. Assuming structural solvability of $\mathcal{E}$, under the perfect intervention $\mathrm{do}(\boldsymbol{X}_{I_i} = \boldsymbol{\xi}_{I_i})$, there is a unique solution $\boldsymbol{X}^*_{\mathrm{do}(\boldsymbol{X}_{I_i}=\boldsymbol{\xi}_{I_i})}$ to the intervened LEE, for any value of $\boldsymbol{\xi}_{I_i} \in \mathcal{R}_{I_i}$. This defines a function $h_i : \mathcal{R}_{\mathrm{pa}_{\mathcal{M}_\mathcal{E}}(i)} \to \mathcal{R}_i$ given by the $i$'th component $h_i(\boldsymbol{\xi}_{\mathrm{pa}_{\mathcal{M}_\mathcal{E}}(i)}) := (\boldsymbol{X}^*_{\mathrm{do}(\boldsymbol{X}_{I_i}=\boldsymbol{\xi}_{I_i})})_i$. The $i$'th structural equation of the induced SCM $\mathcal{M}_\mathcal{E}$ is defined as:

$$X_i = h_i(\boldsymbol{X}_{\mathrm{pa}_{\mathcal{M}_\mathcal{E}}(i)}).$$

---

[4] In Pearl (2000)'s words: "Mathematically, the distinction between structural and algebraic equations is that the latter are characterized by the set of solutions to the entire system of equations, whereas the former are characterized by the solutions of each individual equation. The implication is that any subset of structural equations is, in itself, a valid model of reality—one that prevails under some set of interventions."

Note that this equation is *equivalent* to the labeled equation $\mathcal{E}_i$ in the sense that they have identical solution sets $\{(X_i^*, \boldsymbol{X}^*_{\mathrm{pa}_{\mathcal{M}_\mathcal{E}}(i)})\}$. Repeating this procedure for all $i \in \mathcal{I}$, we obtain the induced SCM $\mathcal{M}_\mathcal{E}$.

This construction is designed to preserve the important mathematical structure. In particular:

**Lemma 1** *Let $\mathcal{E}$ be an LEE, $I \subseteq \mathcal{I}$ and $\boldsymbol{\xi}_I \in \mathcal{R}_I$ and consider the perfect intervention $\mathrm{do}(\boldsymbol{X}_I = \boldsymbol{\xi}_I)$. Suppose that both the LEE $\mathcal{E}$ and the intervened LEE $\mathcal{E}_{\mathrm{do}(\boldsymbol{X}_I=\boldsymbol{\xi}_I)}$ are structurally solvable. (i) Applying the intervention $\mathrm{do}(\boldsymbol{X}_I = \boldsymbol{\xi}_I)$ to the induced SCM $\mathcal{M}_\mathcal{E}$ gives the same result as constructing the SCM corresponding to the intervened LEE $\mathcal{E}_{\mathrm{do}(\boldsymbol{X}_I=\boldsymbol{\xi}_I)}$:*

$$(\mathcal{M}_\mathcal{E})_{\mathrm{do}(\boldsymbol{X}_I=\boldsymbol{\xi}_I)} = \mathcal{M}_{\mathcal{E}_{\mathrm{do}(\boldsymbol{X}_I=\boldsymbol{\xi}_I)}}.$$

*(ii) Solvability of the intervened LEE $\mathcal{E}_{\mathrm{do}(\boldsymbol{X}_I=\boldsymbol{\xi}_I)}$ implies solvability of the induced intervened SCM $\mathcal{M}_{\mathcal{E}_{\mathrm{do}(\boldsymbol{X}_I=\boldsymbol{\xi}_I)}}$ and their respective solutions $\boldsymbol{X}^*_{\mathrm{do}(\boldsymbol{X}_I=\boldsymbol{\xi}_I)}$ are identical.*

**Proof.** The first statement directly follows from the construction of the induced SCM. The key observation regarding solvability is the following. From the construction above it directly follows that $\forall i \in \mathcal{I}$:

$$\forall \boldsymbol{X}_{\mathrm{pa}_\mathcal{E}(i)} \in \mathcal{R}_{\mathrm{pa}_\mathcal{E}(i)} :$$
$$0 = g_i(\boldsymbol{X}_{\mathrm{pa}_\mathcal{E}(i)}) \iff X_i = h_i(\boldsymbol{X}_{\mathrm{pa}_\mathcal{E}(i)\setminus\{i\}}).$$

This trivially implies:

$$\forall \boldsymbol{X} \in \mathcal{R}_\mathcal{I} \forall i \in \mathcal{I} : 0 = g_i(\boldsymbol{X}_{\mathrm{pa}_\mathcal{E}(i)}) \iff X_i = h_i(\boldsymbol{X}_{\mathrm{pa}_{\mathcal{M}_\mathcal{E}}(i)}).$$

This means that each simultaneous solution of all labeled equations is a simultaneous solution of all structural equations, and vice versa:

$$\forall \boldsymbol{X} \in \mathcal{R}_\mathcal{I} : \quad \left( [\forall i \in \mathcal{I} : 0 = g_i(\boldsymbol{X}_{\mathrm{pa}_\mathcal{E}(i)})] \right.$$
$$\left. \iff [\forall i \in \mathcal{I} : X_i = h_i(\boldsymbol{X}_{\mathrm{pa}_{\mathcal{M}_\mathcal{E}}(i)})] \right).$$

The crucial point is that this still holds if an intervention replaces some of the equations (by $0 = X_i - \xi_i$ and $X_i = \xi_i$, respectively, for all $i \in I$). □

## 4.5 From ODEs to deterministic SCMs

We can now combine all the results and definitions so far to construct a deterministic SCM from an ODE under certain stability conditions. We define:

**Definition 10** *An ODE $\mathcal{D}$ is called* structurally stable *if for each $i \in \mathcal{I}$, the ODE $\mathcal{D}$ is stable with respect to $\{\mathcal{I} \setminus \{i\}\}_{i \in \mathcal{I}}$.*

Consider the diagram in Figure 4. Here, the labels of the arrows correspond with the numbers of the sections that discuss the corresponding mapping. The downward mappings correspond with a particular intervention $\mathrm{do}(\boldsymbol{X}_I = \boldsymbol{\xi}_I)$, applied at the different levels (ODE, induced LEE, induced SCM). Our main result:

**Theorem 2** *If both the ODE $\mathcal{D}$ and the intervened ODE $\mathcal{D}_{\mathrm{do}(\boldsymbol{X}_I=\boldsymbol{\xi}_I)}$ are structurally stable, then: (i) The diagram in Figure 4 commutes.[5] (ii) If furthermore, the intervened ODE $\mathcal{D}_{\mathrm{do}(\boldsymbol{X}_I=\boldsymbol{\xi}_I)}$ is stable, the induced intervened SCM $\mathcal{M}_{\mathcal{E}_{\mathcal{D}_{\mathrm{do}(\boldsymbol{X}_I=\boldsymbol{\xi}_I)}}}$ has a unique solution that coincides with the stable equilibrium of the intervened ODE $\mathcal{D}_{\mathrm{do}(\boldsymbol{X}_I=\boldsymbol{\xi}_I)}$.*

**Proof.** Immediate from Theorem 1 and Lemma 1. □

Note that even though the ODE may contain self-loops (i.e., the time derivative $\dot{X}_i$ could depend on $X_i$ itself, and hence $i \in \mathrm{pa}_\mathcal{D}(i)$), the induced SCM $\mathcal{M}_{\mathcal{E}_\mathcal{D}}$ does *not* contain self-loops by construction (i.e., $i \notin \mathrm{pa}_{\mathcal{M}_{\mathcal{E}_\mathcal{D}}}(i)$). Somewhat surprisingly, the structural stability conditions actually imply the existence of self-loops (because if $X_i$ would not occur in the equilibrium equation $(\mathcal{E}_\mathcal{D})_i$, its value would be undetermined and hence the equilibrium would not be unique).

Whether one prefers the SCM representation over the LEE representation is mainly a matter of practical considerations: both representations contain all the necessary information to predict the results of arbitrary perfect interventions, and one can easily go from the LEE representation to the SCM representation. One can also easily go in the opposite direction, but this cannot be done in a unique way. For example, one could rewrite each structural equation $X_i = h_i(\boldsymbol{X}_{\mathrm{pa}_\mathcal{M}(i)})$ as the equilibrium equation $0 = h_i(\boldsymbol{X}_{\mathrm{pa}_\mathcal{M}(i)}) - X_i$, but also as the equilibrium equation $0 = h_i^3(\boldsymbol{X}_{\mathrm{pa}_\mathcal{M}(i)}) - X_i^3$ (in both cases, it would be given the label $i$).

In case the dynamics contains no directed cycles (not considering self-loops), the advantage of the SCM representation is that it is more explicit. Starting at the variables without parents, and following the topological ordering of the corresponding directed acyclic graph, we directly obtain the solution of an SCM by simple substitution in a finite number of steps. When using the LEE representation, one needs to solve a set of equations instead. In the cyclic case, one needs to solve a set of equations in both representations, and the difference is merely cosmetical. However, one could argue that the LEE representation is slightly more natural in the cyclic case, as it does not force us to make additional (structural) stability assumptions.

---

[5]This means that it does not matter in which direction one follows the arrows, the end result will be the same.

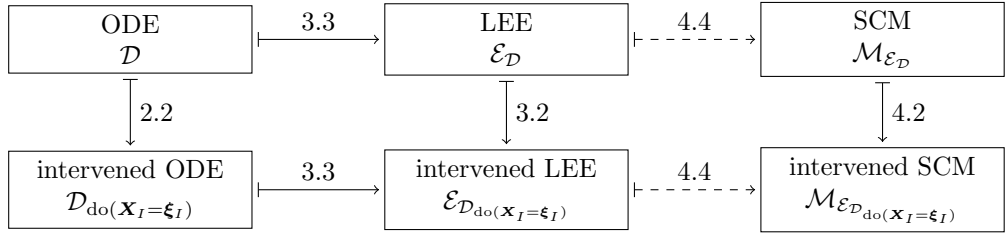

Figure 4: Each of the arrows in the diagram corresponds with a mapping that is described in the section that the label refers to. The dashed arrows are only defined under structural solvability assumptions on the LEE (or structural stability assumptions on the inducing ODE). If the ODE $\mathcal{D}$ and intervened ODE $\mathcal{D}_{\mathrm{do}(\boldsymbol{X}_I=\boldsymbol{\xi}_I)}$ are structurally stable, this diagram commutes (cf. Theorem 2).

### 4.5.1 Example: damped harmonic oscillators

Figure 5 shows the graph of the structural causal model induced by our construction. It reflects the intuition that at equilibrium, (the position of) each mass has a direct causal influence on (the positions of) its neighbors. Observing that the momentum variables always vanish at equilibrium (even for any perfect intervention that we consider), we can decide that the only relevant variables for the SCM are the position variables $Q_i$. Then, we end up with the following structural equations:

$$Q_i = \frac{k_i(Q_{i+1} - l_i) + k_{i-1}(Q_{i-1} + l_{i-1})}{k_i + k_{i+1}}. \qquad (13)$$

## 5 Discussion

In many empirical sciences (physics, chemistry, biology, etc.) and in engineering, differential equations are a common modeling tool. When estimating system characteristics from data, they are especially useful if measurements can be done on the relevant time scale. If equilibration time scales become too small with respect to the temporal resolution of measurements, however, the more natural representation may be in terms of structural causal models. The main contribution of this work is to build an explicit bridge from the world of differential equations to the world of causal models Our hope is that this may aid in broadening the impact of causal modeling.

Note that information is lost when going from a dynamical system representation to an equilibrium representation (either LEE or SCM), in particular the rate of convergence toward equilibrium. If time-series data is available, the most natural representation may be the dynamical system representation. If only snapshot data or equilibrium data is available, the dynamical system representation can be considered to be overly complicated, and one may use the LEE or SCM representation instead.

We have shown one particular way in which structural causal models can be "derived". We do not claim that this is the only way: on the contrary, SCMs can probably be obtained in several other ways and from other representations as well. A recent example is the derivation of SCMs from stochastic differential equations (Sokol and Hansen, 2013). Other related work on differential equations and causality is (Voortman et al., 2010; Iwasaki and Simon, 1994).

We intend to extend the basic framework described here towards the more general stochastic case. Uncertainty or "noise" can enter in different ways: via uncertainty about (constant) parameters of the differential equations, via uncertainty about the initial condition (in the case of constants of motion) and via latent variables (in the case of confounding).

### Acknowledgements

We thank Bram Thijssen, Tom Claassen, Tom Heskes and Tjeerd Dijkstra for stimulating discussions. JM was supported by NWO, the Netherlands Organization for Scientific Research (VENI grant 639.031.036).

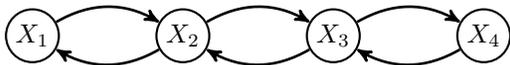

Figure 5: Graph of the structural causal model induced by the mass-spring system for $D = 4$.